\pdfoutput=1

\documentclass[11pt]{article}

\usepackage{acl}

\usepackage{times,amsmath,graphicx,amsthm,amssymb}
\PassOptionsToPackage{table}{xcolor}
\usepackage{mathtools}
\usepackage{latexsym}
\usepackage{multirow}
\usepackage{colortbl}
\usepackage{tabularx}
\usepackage{booktabs}
\usepackage{algorithm}
\usepackage{algpseudocode}

\newcommand{\norm}[1]{\left\lVert#1\right\rVert}
\usepackage[T1]{fontenc}
\usepackage[utf8]{inputenc}
\usepackage{microtype}
\usepackage{inconsolata}
\usepackage{soul}
\usepackage{bbding,pifont}
\newcommand{\cmark}{\ding{51}}%
\newcommand{\xmark}{\ding{55}}%
\title{Overcoming Catastrophic Forgetting by Exemplar Selection in \\
  Task-oriented Dialogue System}

\author{Chen Chen$^1$, Ruizhe Li$^2$, Yuchen Hu$^1$, Yuanyuan Chen$^1$, Chengwei Qin$^{1}$, Qiang Zhang$^3$ \\
$^1$Nanyang Technological University, Singapore
\quad$^2$University of Aberdeen, UK \\
$^3$ Zhejiang University, China \\
}

\begin{document}
\maketitle
\begin{abstract}
  Intelligent task-oriented dialogue systems (ToDs) are expected to continuously acquire new knowledge, also known as Continual Learning (CL), which is crucial to fit ever-changing user needs. However, catastrophic forgetting dramatically degrades the model performance in face of a long streamed curriculum. In this paper, we aim to overcome the forgetting problem in ToDs and propose a method (HESIT) with hyper-gradient-based exemplar strategy, which samples influential exemplars for periodic retraining. Instead of unilaterally observing data or models, HESIT adopts a profound exemplar selection strategy that considers the general performance of the trained model when selecting exemplars for each task domain. Specifically, HESIT analyzes the training data influence by tracing their hyper-gradient in the optimization process. Furthermore, HESIT avoids estimating Hessian to make it compatible for ToDs with a large pre-trained model. Experimental results show that HESIT effectively alleviates catastrophic forgetting by exemplar selection, and achieves state-of-the-art performance on the largest CL benchmark of ToDs in terms of all metrics.
\end{abstract}
\section{Introduction}
Serving as a core technique of smart assistants, task-oriented dialogue systems (ToDs) are expected to continuously acquire new knowledge through time regarding user needs~\cite{madotto2020continual}, e.g., adding fresh slot-value pairs or handling dissimilar tasks. This ability is also known as Continual Learning (CL)~\cite{mundt2020wholistic}, which has recently attracted a surge of interest in NLP community~\cite{biesialska2020continual}, as well as other machine learning techniques~\cite{ qu2021recent}.\par
In this setting, the central problem of CL is \emph{catastrophic forgetting}~\cite{kirkpatrick2017overcoming}, as the data is streamed and the neural model inevitably forgets previously learned knowledge when fitting new training data in the sequential order. Such a phenomenon is particularly conspicuous in ToDs tasks due to the obvious distributional shift between task scenarios, e.g., setting an alarm clock and booking a flight are completely irrelevant tasks. As a result, it is challenging for a single model that is learned from a unidirectional curriculum to simultaneously handle multi-domain tasks.\par

\begin{figure}[t]
  \centering
  \includegraphics[width=0.48\textwidth]{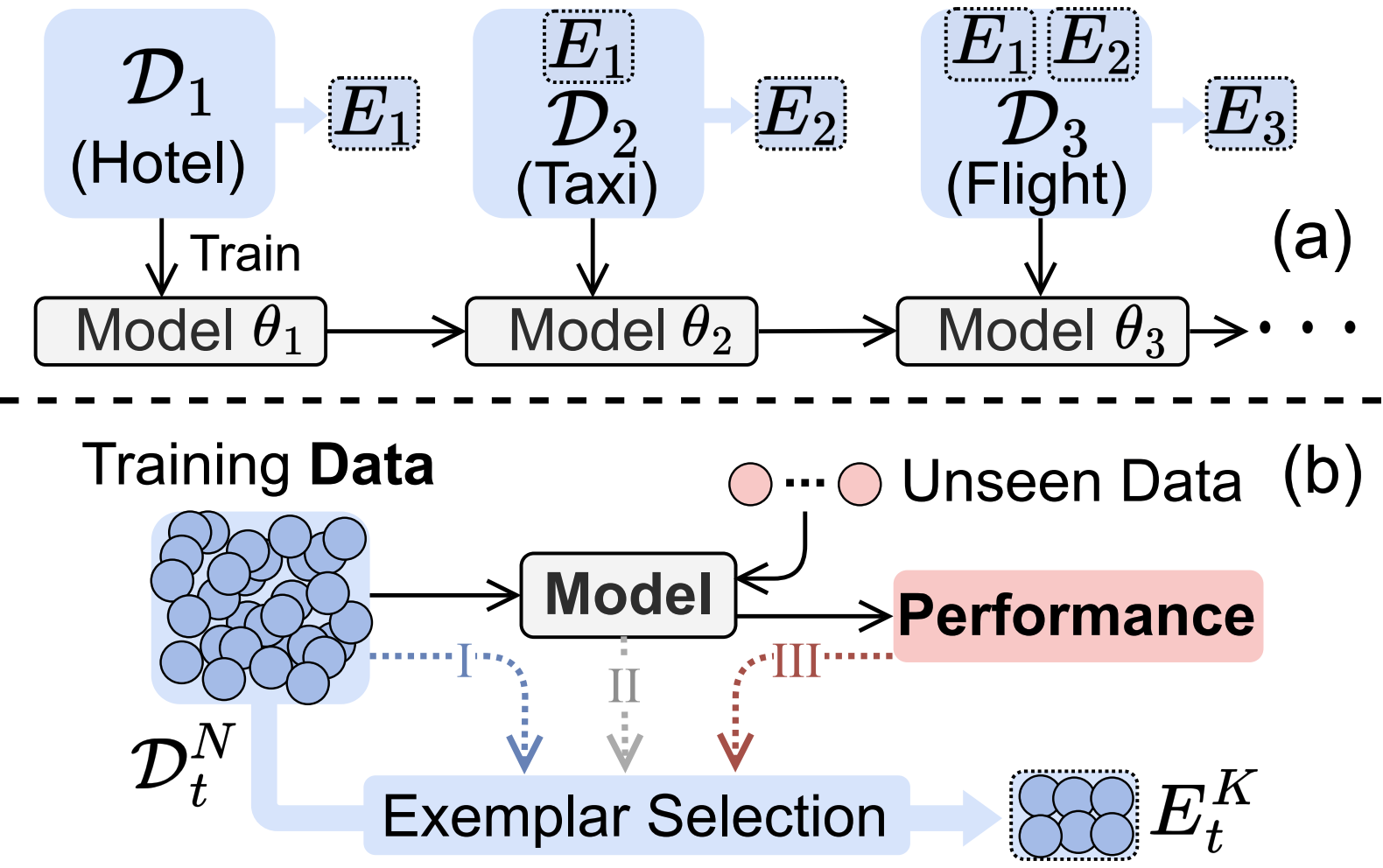}
  \caption{(a) Rehearsal-based CL in Task-oriented dialogue system. Exemplars $E_t$ are sampled from $t$-domain training data for episodic rehearsal. (b) Exemplar selection in terms of influence chain ``\emph{Data} (\uppercase\expandafter{\romannumeral1}) -- \emph{Model} (\uppercase\expandafter{\romannumeral2}) -- \emph{Performance} (\uppercase\expandafter{\romannumeral3})''. Our method penetrates into the performance perspective.}
  \label{fig1}
\end{figure}
To mitigate the problem of forgetting, existing efforts~\cite{de2021continual} follow the three lines: (i) adding regularization terms to consolidate learned knowledge, (ii) developing dynamic architectures for the task-specific domain, and (iii) applying historical data rehearsal. Previous works~\cite{madotto2020continual} have demonstrated that regularization methods gradually lose effectiveness if faced with a long curriculum, and we confirm this belief with a relative experiment. Dynamic architecture methods (e.g. adapter tuning) are usually considered as an \textbf{unfair baseline}~\cite{rusu2016progressive, maltoni2019continuous, mi2020continual} due to multiple sets of parameters, and it requires a further step during inference to determine the correct parameters resulting in high latency for ToDs. Therefore, this work focuses on rehearsal methods (Fig.~\ref{fig1} (a))~\cite{rolnick2019experience} that store a small number of past samples (a.k.a. exemplars) in episodic memory and replay them periodically, which is simple but effective. Then we raise our research question: \emph{which utterances are suitable to be selected as rehearsal exemplars for ToDs?}\par

Although statistic methods (e.g., reservoir algorithm~\cite{chaudhry2019continual}) conduce to select better exemplars than random sampling, according to the accredited influence chain ``\emph{Data-Model-Performance}”~\cite{sun2022exploring}, an exemplar selection strategy should further consider the general performance of models during testing. In other words, we should select the exemplars that positively influence the \emph{performance} of the trained network on unseen data, as shown in Fig.~\ref{fig1} (b).\par

To explore the influence of training data from the performance perspective,
one pioneering work uses the Influence Functions (IFs) based on the derivation chain rule~\cite{koh2017understanding} to quantify the contribution of an individual training sample, by observing its impact on the test loss when removing it. Though IFs have shown effectiveness in image classification tasks, this method requires estimating the inverse Hessian matrix, leading to high computing costs~\cite{guo2020fastif} and unstable results for large pre-trained models that are widely used in ToDs. Furthermore, IFs only measure the contribution around the model parameters at the final training epoch, which fails to trace the data contribution in the dynamic optimization process~\cite{chen2021hydra}. \par
In this paper, we propose a rehearsal-based CL method, named HESIT (Hyper-gradient-based Exemplar Selection by Influence Tracing) to effectively overcome the catastrophic forgetting in ToDs. Specifically, instead of only focusing on the model parameters of the final epoch, HESIT traces and analyzes the hyper-gradient of training examples in the complicated optimization process. More importantly, its exemplar selection strategy roots in model \emph{performance} on unseen data to measure the influence of these traced samples. In this way, influential and representative exemplars can be determined and stored, and then utilized to remind the model of learned knowledge. Moreover, HESIT is Hessian-free to avoid the instability when estimating Hessian matrix of a large pre-trained models, which has been widely used in ToDs. Experimental results demonstrate that HESIT achieves state-of-the-art performance on the largest CL benchmark of ToDs~\cite{madotto2020continual} including 37 domain tasks in terms of all metrics. Furthermore, comparative experiments show that the exemplar selection in HESIT surpasses all other selection strategies from other perspectives.\par
The main contributions of this paper can be summarized as follows:
\begin{itemize}
  \item We propose HESIT -- a rehearsal-based CL method with a novel exemplar selection strategy that can effectively overcome catastrophic forgetting in ToDs.
  \item HESIT dynamically traces the hyper-gradient of candidate data in the training process and selects exemplars in terms of data influence to model performance.
  \item Compared with other influence function-based methods, HESIT avoids estimating the Hessian matrix without degrading effect, which is compatible for ToDs with a large-scale pre-trained model.
  \item Extensive experiments show that HESIT can achieve state-of-the-art performance on the largest CL benchmark of ToDs (37 domain tasks).
\end{itemize}

\section{Related Work}
\noindent\textbf{Continual learning.}
In the past few years, CL has achieved remarkable progress in mitigating catastrophic forgetting~\cite{mccloskey1989catastrophic}, which can be classified into three categories: (1) \emph{regularization-based} methods ~\cite{kirkpatrick2017overcoming, serra2018overcoming,huang2021continual,qin-joty-2022-continual,qin2024lifelong}, which focus on regularizing the parameters corresponding to be inherited from the old tasks and penalizing the feature drift. (2) \emph{architecture-based} methods~\cite{xu2018reinforced,li2019learn,ermis2022continual,qin-etal-2023-lifelong}, which develop dynamic parameter isolation or expansion during CL training, where each task domain learns a specific architecture. (3) \emph{rehearsal-based} methods~\cite{cui2019class,chaudhry2019continual,verwimp2021rehearsal,qin2022lfpt}, which utilize a small replay buffer to retain a fraction of learned training data and utilize them to retain the task knowledge. One shortcoming of architecture-based and rehearsal-based methods is that they require extra memory space for storing parameters and old data samples~\cite{de2021continual}, while regularization-based methods might be overwhelmed when handling many task domains~\cite{mai2022online}. \par
\noindent\textbf{Continual learning in ToDs.}
Early work on CL for ToDs was introduced by~\newcite{lee2017toward}, where  the elastic weight consolidation (EWC) method was utilized to alleviate catastrophic forgetting on 3 sequential domains.~\newcite{wu2019transferable} studied CL for DST sub-task on the MultiWOZ dataset, where several baselines have been compared. ~\newcite{mi2020continual} combined data rehearsal and EWC on NLG sub-task and expanded the learned knowledge to 13 task domains. Furthermore,~\newcite{madotto2020continual} developed a large benchmark for all INTENT, DST, and NLG tasks, where mainstream CL methods (e.g., L2, AGEM, and Adapter) were compared on a total of 37 domains. However, it shows that regularization-based methods lose effectiveness when ToDs encounter dozens of tasks~\cite{madotto2020continual}. Meanwhile, architecture-based methods require a further step to determine which set of parameters to use~\cite{wortsman2020supermasks}, thus blocking the real-time response from ToDs. In this work, we focus on rehearsal-based CL methods as it is simple and effective, with an acceptable cost of extra memory. \par
\noindent\textbf{Exemplar selection.} To sample representative or informative exemplars from a large dataset, reservoir sampling has been introduced in~\cite{isele2018selective,chaudhry2019continual} so that the data distribution in the buffer follows the data distribution that has already been seen.~\newcite{rebuffi2017icarl} proposed a herding-based strategy to maintain an online coreset. Similarly, gradient-based methods have also been proposed in~\cite{broderick2013streaming, aljundi2019gradient} to maximize the variance in the buffer. For ToDs with clear task boundaries,~\newcite{mi2020continual} defined representative exemplars as a small set of utterances that fulfill a loss-based criterion on the current domain. The above-mentioned methods leverage models to choose high-quality data points, but fail to provide an insightful perspective on data contribution analysis in terms of model performance on unseen data. \par
\section{Methodology}
\subsection{Background: End-to-end ToDs} \label{3.1}
We formulate task-oriented dialogue systems as a sequence-to-sequence generation problem that generates both API-calls and system responses. The API-call consists of the user intent and the current dialogue state, which can be empty or the system speech-act, to generate the system response. Thanks to the recent advance in ToDs, all ground-truth information at each turn is provided by existing annotated dialogue datasets. In this setting, the data format of API-call $\mathcal{C}$ is shown as follows:
\begin{equation}
  \mathcal{C}(\mathbf{H})=\underbrace{\mathbf{I}}_\text{Intent} \underbrace{(s_1=v_1, \dots, s_p=v_p)}_\text{Slot-value \ pairs},
  \label{data format}
\end{equation}
where $\mathbf{H}$ is the dialogue history, $\mathbf{I}$ denotes the user intent, and ($s_i$, $v_i$) stands for a slot-value pair from dialogue state tracking. \par
ToDs can be typically decomposed into different modules, including intent recognition (INTENT), history state tracking (DST), and natural language generation (NLG). In this paper, besides the sub-modules, we define the end-to-end (E2E) manner that directly generates the system response $\mathbf{R}$:
\begin{align}
  \mathbf{H} + \mathcal{C}(\mathbf{H})\rightarrow\mathbf{R}
\end{align}
where $\mathcal{C}(\mathbf{H})$ is often empty, and thus the model can directly generate a response according to dialogue history ($\mathbf{H} \rightarrow \mathbf{R}$). In addition, our method HESIT can work for both E2E and modular ToDs. More details are attached in Appendix~\ref{data_format}.

\subsection{Rehearsal-based CL in ToDs}
Given the training data $\mathcal{D}$ consisting of $T$ ToDs task domains $\mathcal{D} = \{\mathcal{D}_1 \cdots \mathcal{D}_T$\}, continual learning aims to train a neural model $f_\theta$ on $\mathcal{D}$ in a sequence of tasks. In each task $t$, new data $\mathcal{D}_t$ is used to update model $f_{\theta_{t-1}}$, while the updated model $f_{\theta_{t}}$ needs to perform well on all tasks so far. In general, the crux of learning $f$ is to overcome the catastrophic forgetting problem where learned knowledge is continually overwritten by streamed data. \par
To suppress it, the rehearsal-based CL methods construct a small size of memory buffer $\mathcal{M}_t$ from each task domain $\mathcal{D}_t$. When training task $t$, the data in $\mathcal{M}_{1:{t-1}}$ will be retrained together with $\mathcal{D}_t$. Accordingly, a specific training step of task $t$ is written as:
\begin{equation}
  \begin{split}
    &\hat{\theta} = \arg \min _{\theta}\frac{1}{N}\sum_{(x_i,y_i) \in \mathcal{D}_t^\textrm{trn}}\ell( (x_{i}, y_{i}), \theta ) \\
    \textrm{s.t.} \quad & (x_{i}, y_{i}) \in \mathcal{M}_{1:{t-1}}\cup\mathcal{D}_t^\textrm{trn}
  \end{split}
\end{equation}
where $N$ is the total number of training data points from $T$ tasks, $\ell$ denotes the empirical loss and $(x_{i}, y_{i})$ is a training data point of task $t$. Considering the storage burden of $\mathcal{M}$, the retained samples, called exemplars, should be informative or influential for $\mathcal{D}_t^\textrm{trn}$.

\subsection{HESIT: Exemplar Selection}
In this subsection, we present the details of the exemplar selection strategy to find influential utterances in $\mathcal{M}_t$, which also serves as the core technique in HESIT.\par
Intuitively, given a training sample $z_i \coloneqq (x_i,y_i) \in \mathcal{D}_t^\textrm{trn}$, we want to quantify its influence on the validation loss $\mathcal{L}(\mathcal{D}_t^\textrm{val},\hat{\theta})$ via the resulting model $\hat{\theta}$. One possible way is to find out the loss difference $\mathcal{I}(z_i,\mathcal{D}_t^\textrm{val}) \coloneqq \mathcal{L}(\mathcal{D}_t^\textrm{val},\hat{\theta}_{\bar{z}_i})-\mathcal{L}(\mathcal{D}_t^\textrm{val},\hat{\theta})$ where $\hat{\theta}_{\bar{z}_i}$ would be the resulting model if $z_i$ were not present in $\mathcal{D}_t^\textrm{trn}$. For this purpose, we introduce a weight variable $\epsilon_i$ and consider the new full-set training loss:
\begin{equation}
  \mathcal{L}_\textrm{trn}(\theta) = \frac{1}{N} \sum_{(x_i,y_i) \in \mathcal{D}_t^\textrm{trn}} ( \ell((x_i,y_i),\theta)+\epsilon_i \ell(z_i,\theta)).
\end{equation}
Now $\hat{\theta}$ and $\mathcal{I}(z_i,\mathcal{D}_t^\textrm{val})$ are taken as functions of $\epsilon_i$ and $\mathcal{I}(z_i,\mathcal{D}_t^\textrm{val})$ can be approximated by the first order Taylor expansion:
\begin{equation}
  \begin{split}
    \mathcal{I}(z_i,\mathcal{D}_t^\textrm{val}) & \approx -\frac{1}{N} \frac{\mathrm{d}\mathcal{L}(\mathcal{D}_t^\textrm{val},\hat{\theta})}{\mathrm{d}\epsilon_i} {|}_{\epsilon_i = 0} \\
    & = -\frac{1}{N} \frac{\partial \mathcal{L}(\mathcal{D}_t^\textrm{val},\hat{\theta})}{\partial \hat{\theta}}\frac{\mathrm{d}\hat{\theta}}{\mathrm{d}\epsilon_i} {|}_{\epsilon_i = 0}.
    \label{cij}
  \end{split}
\end{equation}
While $\frac{\partial \mathcal{L}(\mathcal{D}_t^\textrm{val},\hat{\theta})}{\partial \hat{\theta}}$ can be exactly calculated using gradient back-propagation on the validation data, computing $\nabla_i \coloneqq \frac{\mathrm{d}\hat{\theta}}{\mathrm{d}\epsilon_i}$, which is also named hyper-gradient~\cite{bengio2000continuous}, is not straightforward because the parameter $\epsilon_i$ is involved in the entire optimization process. \par
In gradient descent based optimization, for each iteration step $r$, we have $\theta_r = \theta_{r-1} - \gamma g_{r-1}$ where $\gamma$ is the learning rate and $g_{r-1}$ is the batch gradient. Then $\nabla_i$ can be computed recursively as:
\begin{equation}
  \nabla_{i,r} = \nabla_{i,r-1}-\gamma (H_{r-1} \nabla_{i,r-1} + \frac{\partial g_{r-1}}{\partial \epsilon_i})
\end{equation}
where $H_{r-1}$ denotes the Hessian of the batch loss with respect to $\theta_{r-1}$. Most related works impose some assumptions on the model to ensure that $H$ is invertible and attempt to approach the inverse Hessian vector product by numerical methods~\cite{koh2017understanding}. However, the assumptions are too strong to be feasible in practice, and with increasing model and dataset size, the numerical methods are time-consuming and may lead to diverging results.~\citet{chen2021hydra} propose a faster scheme by discarding the Hessian computations, which have a bounded error. Therefore, HESIT can recurrently update the $\nabla_{i,r}$ through the optimization trajectory with an acceptable time cost. We further design a validation experiment in Section~\ref{exp3} to demonstrate that such a manner can successfully estimate data influence without the estimation of Hessian. \par
\begin{algorithm}[t!]
  \caption{Influential Example Tracing}
  \begin{algorithmic}[1]
    \Require
    Traced example set $\mathcal{Z}_t = \{z_i\}_{i=1}^I$ of task $t$, training dataset $\mathcal{D}^\textrm{trn}_t$, validation dataset $\mathcal{D}^\textrm{val}_t$, batch size $B$, training dataset size $N$, total iteration steps $R$, learning rate $\gamma(t)$
    \Ensure
    Model $\hat{\theta}$, influence values $\mathcal{I}(\mathcal{Z},\mathcal{D}_t^\textrm{val})$
    \State Setup reproducible training environment
    \State Train model with $\mathcal{D}^\textrm{trn}_t$ and obtain $\hat{\theta}$
    \State $\boldsymbol{v} \leftarrow \frac{\partial \mathcal{L}(\mathcal{D}_t^\textrm{val},\hat{\theta})}{\partial \hat{\theta}}$
    \State Reset training environment and retrain model with $\mathcal{D}^\textrm{trn}_t$ for second identical training
    \For{$i=1$ to $I$}
    \State $\boldsymbol{v} \nabla_{i,0} \leftarrow 0$
    \EndFor
    \For{$r=1$ to $R$}
    \If{$\mathrm{current\ batch\ contains}\ z_i$}
    \State $\boldsymbol{v} \nabla_{i,r} \leftarrow \gamma(t)\boldsymbol{v} \nabla_{i,r-1} - \frac{N}{B}\boldsymbol{v} g_{i,r-1}$
    \Else
    \State $\boldsymbol{v} \nabla_{i,r} \leftarrow \gamma(t)\boldsymbol{v} \nabla_{i,r-1}$
    \EndIf
    \EndFor
    \For{$i=1$ to $I$}
    \State $ \mathcal{I}(z_i,\mathcal{D}^\textrm{val}_t) \leftarrow -\frac{1}{N} \boldsymbol{v} \nabla_{i,R} $
    \EndFor
  \end{algorithmic}
  \label{algo1}
\end{algorithm}
\begin{algorithm}[t]
  \caption{HESIT Training}
  \begin{algorithmic}[1]
    \Require
    Sequential $T$-domain training data $\mathcal{D}=\{(\mathcal{D}_1^\textrm{trn},\mathcal{D}_1^\textrm{val}),\ldots,(\mathcal{D}_T^\textrm{trn}, \mathcal{D}_T^\textrm{val})\}$
    \Ensure
    Parameter $\theta_T$ that handles T-domain tasks
    \State Initialize the model parameter $\theta_0$ and replay buffer $\mathcal{M}_0$
    \For{$t=1$ to $T$}
    \State Sample $\mathcal{Z}_t\sim \mathcal{D}_t^\textrm{trn}$
    \State Calculate $\mathcal{I}(\mathcal{Z}_t,\mathcal{D}_t^\textrm{val})$ and $\theta_{t-1}$ by Algo.\ref{algo1}
    \State Select $E^K_t \subset \mathcal{Z}_t$ by top $\norm{\mathcal{I}(\mathcal{Z}_t,\mathcal{D}_t^\textrm{val})}$
    \While{\emph{not converge}}
    \State Update $\theta_{t-1}$ using \{$\mathcal{D}_t^\textrm{trn}\cup \mathcal{M}_{t-1}\}$
    \EndWhile
    \State $\mathcal{M}_t \leftarrow E^K_t\cup \mathcal{M}_{t-1}$
    \EndFor
  \end{algorithmic}
  \label{algo2}
\end{algorithm}
As illustrated in Algorithm~\ref{algo1}, the influence scores of training examples are computed via reproducible retraining to reduce the space complexity. In this way, the time complexity to trace hyper-gradient is $\mathcal{O}(I\cdot R\cdot \omega)$, where $\omega$ is the time spent in computing parameter gradient, but the space complexity has been reduced to $\mathcal{O}(I)$. Finally, a set of exemplars $E^K_t$ are picked out according to the influence scores, with top-$K$ strategy.

\subsection{HESIT: Training Schedule}
We present the training schedule of HESIT in Algorithm~\ref{algo2}, which integrates the exemplar selection strategy and rehearsal-based CL into ToDs. Specifically, for each domain $t$, we select $E_t^K$ from $\mathcal{D}_t$ using a hyper-gradient based strategy and feed them into the replay buffer $\mathcal{M}$. Then $\mathcal{M}$ will be involved in the subsequent training process for retraining the neural model. \par
To further accelerate training, we adopt two simplifications. Firstly, instead of tracing all training examples, we sample and trace a subset $\mathcal{Z}_t \gg E_t^K$ from each $\mathcal{D}^\textrm{trn}_t$ to reduce the cost. Secondly, tracing is performed only in the first $R$ iterations rather than the entire training process, as the main benefits of optimization have been obtained in these iterations. In addition, the subsequent training (line 6-8) is depend on the size of domain $t$. \par

\section{Experiment}
\subsection{Dataset}
To evaluate the performance of our method, we employ the largest CL benchmark of the ToDs task developed in~\cite{madotto2020continual}. Specifically, four ToD datasets are merged: TaskMaster 2019 (TM19)~\cite{byrne2019taskmaster}, TaskMaster 2020 (TM20)~\cite{byrne2019taskmaster}, Schema Guided Dialogue (SGD)~\cite{lin2021bitod}, and MultiWoZ~\cite{budzianowski2018multiwoz}. Consequently, a curriculum of 37 domains is to be learned continuously in ToDs, where three modules (INTENT, DST, and NLG) are well annotated after pre-processing. We consider two settings which are \textbf{Modularized setting} that learns three modules separately, and \textbf{E2E setting} that learns these modules in a unified manner, as illustrated in~\ref{3.1}. \par
In addition, we summarize the main statistics and detailed sample numbers for each domain in Appendix~\ref{dataset}. It is noted that the amount of examples is highly imbalanced across different domains, which ranges from a few hundred to more than 36k. This distribution is more proximate to reality, as some task domains lack training data in practice.
\subsection{Evaluation Metrics}
Based on three modules, we employ the following well-defined metrics to evaluate system performance:
\begin{itemize}
  \item INTENT recognition is directly measured in terms of accuracy between the predicted intent and ground-truth intent.
  \item DST is evaluated by Joint Goal Accuracy (JGA)~\cite{wu2019transferable} over the ground-truth dialogue.
  \item NLG is evaluated by two metrics: 1) BLEU score~\cite{papineni2002bleu} which is calculated by the distance between generated response and reference sentence, and 2) slot error rate (EER)~\cite{wen2015semantically} which is computed as the ratio between the total number of slots and the values not appearing in the response.
\end{itemize}
For all metrics except EER, a higher value denotes better performance.
In addition, same as~\cite{kale2020few}, we do not calculate EER count for the SGD dataset, since some slots of the SGD dataset have only binary values, \emph{e.g.,} yes or no, which is unfair to calculate average with others.

\subsection{Baselines}
In order to compare the effects of different CL methods, we investigate both regularization-based and rehearsal-based baselines in the ToDs benchmark. Architecture-based methods are excluded as they require a further decision step to determine the agnostic domain of the test set, resulting in a slow system response.\par
\noindent \textbf{Regularization baseline.} We consider two kinds of regularization terms to constrain the model parameter update, which is identity function (\textbf{L2}) and Fisher information matrix (\textbf{EWC}) proposed in ~\newcite{kirkpatrick2017overcoming}. \par
\noindent \textbf{Rehearsal baseline.} We first employ 2 kinds of mainstream rehearsal-based CL methods, which are \textbf{A-GEM}~\cite{chaudhry2018efficient}, and \textbf{LAMOL}~\cite{sun2019lamol}. Since our contribution focuses on exemplar selection, we further reproduce 4 coreset selection strategies in the ToDs benchmark, where 2 baselines are from data-perspective: (1) \textbf{Random} denotes that exemplars are randomly sampled from the train set, and (2) \textbf{UNIFORM} denotes that we uniformly select exemplars to ensure all user intents would be contained in the buffer. Meanwhile, the other two baselines are from the model perspective: (3) \textbf{GSS}~\cite{aljundi2019gradient} selects the examples that maximize the gradient variance in the replay buffer. (4) \textbf{ARPER}~\cite{mi2020continual} selects exemplars from the training set that obtains minimum loss-based criterion. (1)$\sim$(4) share the same training schedule with the proposed \textbf{HESIT} but they adopt different exemplar selection strategies respectively.  \par
Additionally, we attach a lower-bound baseline, called \textbf{VANILLA}, which is trained on each task continuously without any anti-forgetting mechanism. We also provide the \textbf{MULTI} baseline which trains models with all data in an integrated curriculum simultaneously, which is widely viewed as the upper bound of CL methods.
\subsection{Experimental Setup}
For all experiments, we leverage the pre-trained GPT-2~\cite{radford2019language} as the ToDs backbone. Each domain is trained for 10 epochs with early stopping over the validation set. The learning rate is set as 0.001 with a warm-up schedule. In L2 and EWC, the regularization weight is set as 0.001. Considering the buffer memory, all data replay-based methods sample 50 exemplars for each task domain. In HESIT, the $\mathcal{Z}_t$ is set as 1000 for each domain, and the interested training examples are traced for the first 5 epochs. Task order in the curriculum usually has a slight impact on the final performance. To avoid contingency, all experiments are repeated three times and report the average results.   \par


\begin{table}[t]
\small
  \resizebox{1.0\columnwidth}{!}{
    \begin{tabular}{cc|cccc}
      \toprule[1.5pt]
      \multirow{2}{*}{Method} & \multirow{2}{*}{Mem.} & INTENT                                                                     & DST                   & \multicolumn{2}{c}{NLG}                         \\
                              &                       & \emph{Accuracy} $\uparrow$                                                 & \emph{JGA} $\uparrow$ & \emph{EER}$\downarrow$  & \emph{BLEU}$\uparrow$ \\ \midrule
      \emph{VANILLA}          & $\emptyset$           & 2.65                                                                       & 9.33                  & 50.91                   & 4.49                  \\
      \multicolumn{6}{c}{\cellcolor[HTML]{E0E0E0}Regularization-based methods}                                                                                                                               \\
      \emph{L2}               & $\emptyset$           & 2.33                                                                       & 6.85                  & 56.42                   & 5.08                  \\
      \emph{EWC}              & $\emptyset$           & 2.46                                                                       & 8.98                  & 52.61                   & 4.70                  \\
      \multicolumn{6}{c}{\cellcolor[HTML]{E0E0E0}Rehearsal-based methods}                                                                                                                                    \\
      \emph{A-GEM}            & $t|\mathcal{M}|$      & 31.02                                                                      & 11.23                 & 60.98                   & 4.53                  \\
      \emph{LAMOL}            & $\emptyset$           & 2.68                                                                       & 9.42                  & 66.31                   & 3.82                  \\
                              &                       & \multicolumn{4}{c}{\cellcolor[HTML]{EFEEEC}Data-perspective}                                                                                         \\
      \emph{RANDOM}           & $t|\mathcal{M}|$      & 78.22                                                                      & 29.47                 & 19.36                   & 16.92                 \\
      \emph{UNIFORM}          & $t|\mathcal{M}|$      & 80.67                                                                      & 28.94                 & 19.63                   & 17.84                 \\
                              &                       & \multicolumn{4}{c}{\cellcolor[HTML]{EFEEEC}Model-perspective}                                                                                        \\
      \emph{GSS}              & $t|\mathcal{M}|$      & 81.42                                                                      & 30.33                 & 17.48                   & 17.97                 \\
      \emph{ARPER}            & $t|\mathcal{M}|$      & 77.60                                                                      & 27.82                 & 20.52                   & 16.44                 \\
                              &                       & \multicolumn{4}{c}{\cellcolor[HTML]{EFEEEC}Performance-perspective (ours)}                                                                           \\
      \emph{HESIT}            & $t|\mathcal{M}|$      & \textbf{83.46}                                                             & \textbf{31.22}        & \textbf{16.78}          & \textbf{18.25}        \\ \midrule
      \emph{MULTI}            & -                     & 95.45                                                                      & 48.90                 & 12.56                   & 23.61                 \\ \bottomrule[1.5pt]
    \end{tabular}}
  \caption{\textbf{E2E} results on the test set in terms of INTENT accuracy, JGA, EER, and BLEU. ``Mem.'' denotes the memory size of the buffer, where $t=37$ and $|\mathcal{M}|=50$.}
  \vspace{-0.2cm}
  \label{table1}
\end{table}


\begin{table}[t]
  \resizebox{1.0\columnwidth}{!}{
    \begin{tabular}{cc|cccc}
      \toprule[1.5pt]
      \multirow{2}{*}{Method} & \multirow{2}{*}{Mem.} & INTENT                                                                     & DST                   & \multicolumn{2}{c}{NLG}                         \\
                              &                       & \emph{Accuracy} $\uparrow$                                                 & \emph{JGA} $\uparrow$ & \emph{EER}$\downarrow$  & \emph{BLEU}$\uparrow$ \\ \midrule
      \emph{VANILLA}          & $\emptyset$           & 3.06                                                                       & 10.28                 & 18.09                   & 10.42                 \\
      \multicolumn{6}{c}{\cellcolor[HTML]{E0E0E0}Regularization-based methods}                                                                                                                               \\
      \emph{L2}               & $\emptyset$           & 3.59                                                                       & 9.94                  & 18.16                   & 11.13                 \\
      \emph{EWC}              & $\emptyset$           & 3.72                                                                       & 10.06                 & 18.12                   & 11.70                 \\
      \multicolumn{6}{c}{\cellcolor[HTML]{E0E0E0}Rehearsal-based methods}                                                                                                                                    \\
      \emph{A-GEM}            & $t|\mathcal{M}|$      & 10.57                                                                      & 9.86                  & 36.22                   & 6.40                  \\
      \emph{LAMOL}            & $\emptyset$           & 2.72                                                                       & 9.44                  & 35.83                   & 4.43                  \\
                              &                       & \multicolumn{4}{c}{\cellcolor[HTML]{EFEEEC}Data-perspective}                                                                                         \\
      \emph{RANDOM}           & $t|\mathcal{M}|$      & 79.92                                                                      & 39.54                 & 5.82                    & 21.33                 \\
      \emph{UNIFORM}          & $t|\mathcal{M}|$      & 81.10                                                                      & 39.92                 & 5.42                    & 21.17                 \\
                              &                       & \multicolumn{4}{c}{\cellcolor[HTML]{EFEEEC}Model-perspective}                                                                                        \\
      \emph{GSS}              & $t|\mathcal{M}|$      & 81.36                                                                      & 39.80                 & 5.17                    & \textbf{21.62}        \\
      \emph{ARPER}            & $t|\mathcal{M}|$      & 79.63                                                                      & 39.21                 & 6.08                    & 21.12                 \\
                              &                       & \multicolumn{4}{c}{\cellcolor[HTML]{EFEEEC}Performance-perspective (ours)}                                                                           \\
      \emph{HESIT}            & $t|\mathcal{M}|$      & \textbf{82.71}                                                             & \textbf{40.04}        & \textbf{5.11}           & 21.48                 \\ \midrule
      \emph{MULTI}            & -                     & 87.50                                                                      & 50.03                 & 3.42                    & 26.15                 \\ \bottomrule[1.5pt]
    \end{tabular}}
  \caption{\textbf{Modularized} results on the test set in terms of INTENT accuracy, JGA, EER, and BLEU.}
  \vspace{-0.2cm}
  \label{table2}
\end{table}

\section{Result and Analysis}
\subsection{Main Comparison Results}
\noindent \textbf{E2E setting.} We report the main E2E results on the full test set in Table~\ref{table1}, where all metrics are evaluated at the end of the curriculum (37 task domains). We observe that regularization baselines (L2 and EWC) can not alleviate catastrophic forgetting well, and even achieve worse performance than VANILLA in terms of most metrics. The learning curve of EWC is visualized in Fig.~\ref{fig2}. The reason is that the regularization item gradually constrains the optimization space of the neural model, making it difficult to handle subsequent tasks in a long curriculum. Similarly, due to clear task boundaries of ToDs, A-GEM and LAMOL baselines also lose effectiveness as their learning schedules depend on correlations across different domains. In this case, regardless of the selection strategy, storing some raw data and retraining the model periodically can obtain remarkable performance gain. As a performance-based selection strategy, our proposed HESIT  performs significantly better than the RANDOM baseline, which respectively achieves 6.70\%, 5.94\%, 13.33\%, and 7.86\% relative improvement in terms of INTENT accuracy, JGA, EER, and BLEU. In addition, compared with the model perspective GSS baseline, HESIT respectively achieves 2.04\%, 2.93\%, 4.0\%, and 1.48\% relative improvement.\par

\noindent \textbf{Modularized setting.} We summarize modularized results in Table~\ref{table2} that INTENT, DST, and NLG modules are evaluated separately. Compared with the E2E setting, we observe some result fluctuations in different evaluation metrics. On the one hand, each neural model focuses on a single module leading to performance gains. Especially in the NLG sub-task, the modularized model surpasses the E2E model by a large margin, where the EER of MULTI reduces nearly fourfold (12.56\%$\rightarrow$3.42\%). On the other hand, the modularized model ignores the relevance of each sub-task, resulting in performance degradation according to INTENT accuracy (95.45\%$\rightarrow$87.50\%). Regardless of E2E or modularized settings, HESIT shows superiority in alleviating catastrophic forgetting in most metrics. \par

\begin{figure}[tb]
  \centering
  \includegraphics[width=0.49\textwidth]{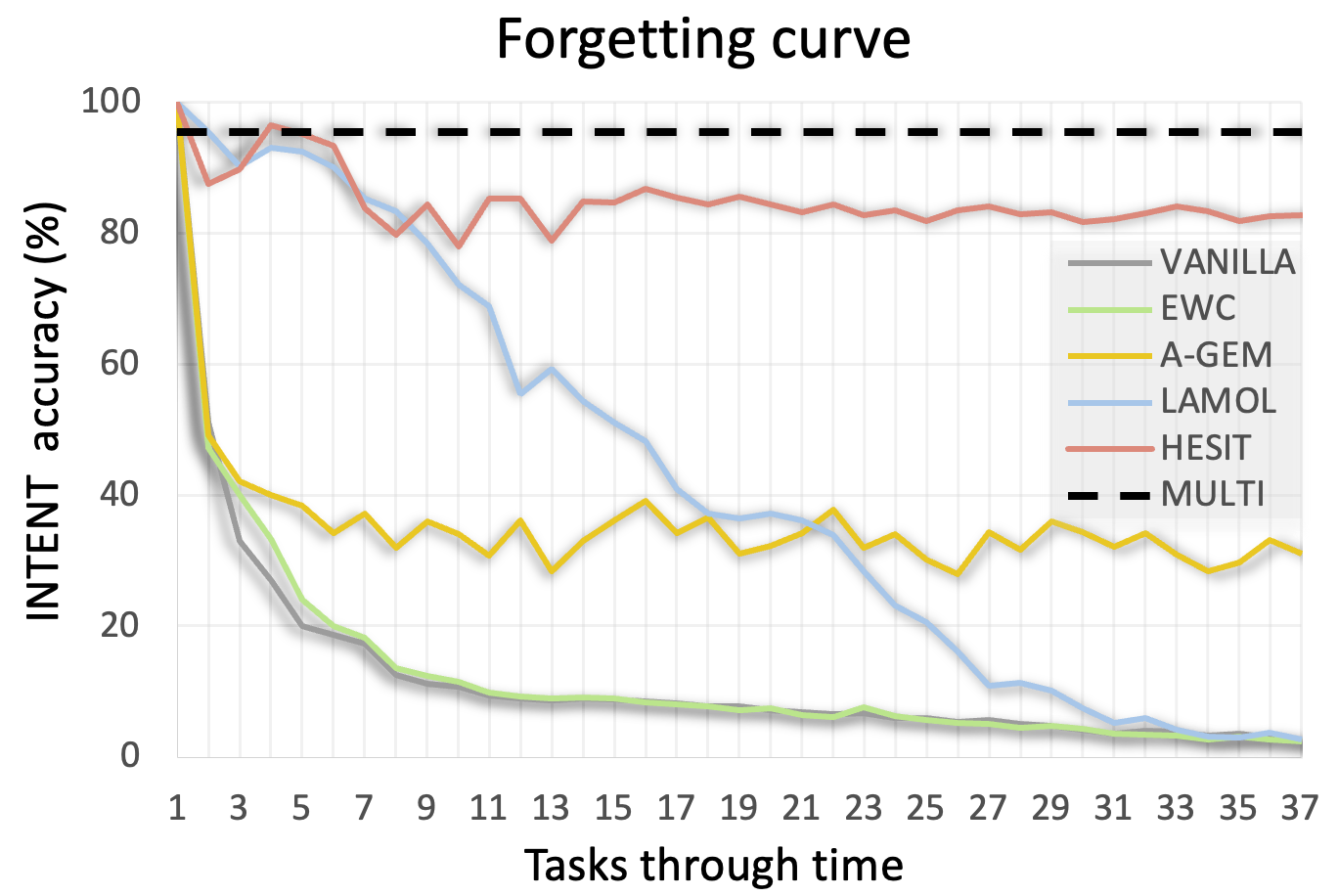}
 \vspace{-0.3cm}
  \caption{Learning curve for INTENT accuracy in E2E setting. Each test point is evaluated on the already learned task in the curriculum.}
  \vspace{-0.5cm}
  \label{fig4}
\end{figure}

\noindent \textbf{Forgetting curve.} We visualize the learning curve of INTENT accuracy in the 37-domain curriculum to observe how HESIT overcomes catastrophic forgetting. To avoid the overlapping of the chart, partial baselines are selected for comparison and shown in Fig.~\ref{fig4}. It is worth noting that we fix the task order in the curriculum, and each test point on the x-axis evaluates the model performance on the test set of learned tasks. \par
From the forgetting curve, we observe that 1) Catastrophic forgetting happened in the VANILLA baseline, and the model performance obviously degrades over time. 2) EWC, A-GEM, and LAMOL can alleviate catastrophic forgetting in the first few tasks, but gradually fail to handle learned tasks with an increase in task amounts. In addition, LAMOL achieves outstanding accuracy at the beginning of the curriculum and performs better than HESIT. 3) For HESIT, there are some fluctuations in performance, but then stabilize at a high performance in terms of accuracy.
\begin{table}[]
  \resizebox{1.0\columnwidth}{!}{
    \begin{tabular}{c|ccccc}
      \toprule[1.5pt]
      \multirow{2}{*}{Method}         & \multirow{2}{*}{\# Er.}                                             & INTENT                     & DST                   & \multicolumn{2}{c}{NLG}                         \\
                                      &                                                                     & \emph{Accuracy} $\uparrow$ & \emph{JGA} $\uparrow$ & \emph{EER}$\downarrow$  & \emph{BLEU}$\uparrow$ \\ \midrule
      \emph{VANILLA}                  & -                                                                   & 2.65                       & 9.33                  & 50.91                   & 4.49                  \\ \midrule
                                      & \multicolumn{5}{c}{\cellcolor[HTML]{EFEEEC}Data-perspective}                                                                                                               \\
      \multirow{4}{*}{\emph{UNIFORM}} & 20                                                                  & 65.15                      & 24.08                 & 23.31                   & 16.47                 \\
                                      & 30                                                                  & 72.22                      & 25.98                 & 22.01                   & 17.19                 \\
                                      & 40                                                                  & 77.56                      & 27.80                 & 19.36                   & 17.44                 \\
                                      & 50                                                                  & 80.67                      & 28.94                 & 19.63                   & 17.84                 \\ \midrule
                                      & \multicolumn{5}{c}{\cellcolor[HTML]{EFEEEC}Model-perspective}                                                                                                              \\
      \multirow{4}{*}{\emph{GSS}}     & 20                                                                  & 66.78                      & 23.85                 & \textbf{18.99}          & \textbf{17.04}        \\
                                      & 30                                                                  & 74.32                      & 26.68                 & 21.02                   & \textbf{17.55}        \\
                                      & 40                                                                  & 78.09                      & 29.76                 & 18.61                   & 17.70                 \\
                                      & 50                                                                  & 81.42                      & 30.33                 & 17.48                   & 17.97                 \\  \midrule
                                      & \multicolumn{5}{c}{\cellcolor[HTML]{EFEEEC}Performance-perspective}                                                                                                        \\
      \multirow{4}{*}{\emph{HESIT}}   & 20                                                                  & \textbf{70.32}             & \textbf{24.97}        & 19.94                   & 16.81                 \\
                                      & 30                                                                  & \textbf{72.85}             & \textbf{27.85}        & \textbf{18.40}          & 17.45                 \\
                                      & 40                                                                  & \textbf{80.01}             & \textbf{30.29}        & \textbf{18.52}          & \textbf{17.93}        \\
                                      & 50                                                                  & \textbf{83.46}             & \textbf{31.22}        & \textbf{16.78}          & \textbf{18.25}        \\  \bottomrule[1.5pt]
    \end{tabular}}
  \caption{E2E results using different exemplar selection strategies. ``\# Er.'' denotes the amount of exemplars for each task domain.}
  \vspace{-0.3cm}
  \label{table3}
\end{table}
\begin{table*}[]
  \centering
  \resizebox{1.99\columnwidth}{!}{
    \begin{tabular}{c|c|c|c|c|c}
      \toprule[1.5pt]
      ID                 & Domain                 & Module                                & Type                                                                  & Utterance                                                                                                                                               & Score \\ \midrule
      \multirow{2}{*}{1} & \multirow{2}{*}{Train} & \multirow{2}{*}{INTENT}               & \multirow{2}{*}{$\mathbf{H}$ $\rightarrow$ $\mathcal{C}(\mathbf{H})$} & \multicolumn{1}{l|}{\multirow{2}{*}{\shortstack[l]{USER: "Yes, I would like a taxi to the Town Centre. From the train station."                                 \\ API: "taxi\_inform() train\_inform() [eos]" } }}  &   \multirow{2}{*}{$-0.64$}   \\
                         &                        &                                       &                                                                       &                                                                                                                                                         &       \\ \midrule
      \multirow{2}{*}{2} & \multirow{2}{*}{Taxi}  & \multirow{2}{*}{\shortstack[c]{INTENT                                                                                                                                                                                                                                           \\ \& DST}}& \multirow{2}{*}{$\mathbf{H}$ $\rightarrow$ $\mathcal{C}(\mathbf{H})$}  & \multirow{2}{*}{\shortstack[l]{USER: "I need the phone number and location of the nearest Red Lobster in the downtown Cambridge area?"         \\ API: "taxi\_request(phone="?") [eos]'' } }  &  \multirow{2}{*}{$-0.77$}  \\
                         &                        &                                       &                                                                       &                                                                                                                                                         &       \\\midrule
      \multirow{2}{*}{3} & \multirow{2}{*}{Hotel} & \multirow{2}{*}{NLG}                  & \multirow{2}{*}{$\mathbf{H}$ $\rightarrow$ $\mathbf{R}$}              & \multicolumn{1}{l|}{\multirow{2}{*}{\shortstack[l]{USER: "Yes, I am attending a physician's conference and need to locate a room for tonight. API-OUT:"         \\ SYSTEM: "Do you have an area or price preference? [eos]" } }}  &  \multirow{2}{*}{$-0.30$}\\
                         &                        &                                       &                                                                       &                                                                                                                                                         &       \\
      \bottomrule[1.5pt]
    \end{tabular}}
  \vspace{-0.25cm}
  \caption{Detrimental examples with low score influences selected by HESIT from the Multi-WoZ dataset. Scores of traced examples are re-normalized to [-1,1].}
  \vspace{-0.3cm}
  \label{table4}
\end{table*}
\subsection{Analysis for Exemplar Selection}
\noindent \textbf{Amount of exemplar.} Using different selection strategies, we vary the size of the memory buffer containing [20,30,40,50] exemplars for each domain. The results are shown in Table~\ref{table3}. We observe that 20 exemplars can significantly alleviate cartographic forgetting, benefiting all metrics compared with the VANILLA baseline. As the buffer size increases, all systems achieve better performance including each ToDs module. Specifically, GSS achieves the best NLG performance when the exemplar amount is 20. In other cases, HESIT performs the best regardless of the exemplar amount, demonstrating its superiority over the performance perspective methods. \par
\noindent \textbf{Case study.} As each training sample is assessed using the Algo.~\ref{algo1}'s Influence Functions to evaluate the impact of unseen data, we conduct a case study on several detrimental examples with low influence scores to examine how HESIT eliminates these negative examples. To this end, we simultaneously trace 1k training points of each domain and randomly sample some low-score examples in Table~\ref{table4}. The first two examples adhere to the format of API-call, and the third example is a system response. In the first two cases, the HESIT successfully identifies the incorrect train\_inform() intent and the missing value-slot pairs with extremely negative scores. The third case is likewise eliminated by the HESIT with a negative score because the model is confused by the empty API-OUT from the history.

\subsection{Effect of HESIT}
\label{exp3}
In this part, we conduct experiments to demonstrate that HESIT can successfully measure the training data influence without estimating Hessian. To this end, we construct two Hessian-involved IFs baselines \textbf{Conjugate gradients (CG)} and \textbf{Stochastic estimation (LISSA)}~\cite{agarwal2017second} and one Hessian-free \textbf{Hydra}~\cite{chen2021hydra} for comparison. As the Hessian of a large pre-trained model is likely irreversible, we further employ a CNN backbone and CIFAR-10 dataset~\cite{krizhevsky2009learning} to evaluate both computation time and performance. The details of the dataset, model and baselines are attached in Appendix~\ref{cifar}.\par
We vary the amount of training and validation data and report the time spent in influence analysis in Table~\ref{table5}. ``Hes.'' indicates whether the method needs to estimate the Hessian of the model. ``Max\_Iter.'' denotes CG iteration in~\cite{martens2010deep}, and is set according to $\mathbf{T}$. In LISSA, ``depth'' and ``repeat'' are two hyper-parameters that influence the computation time, and generally, their products should be equal to $\mathbf{T}$. 
By comparing the baselines and HESIT on ($\mathbf{T}$, $\mathbf{V}$) training-validation dataset pairs of varying sizes, we find that HESIT, Hydra and LISSA require comparable calculation time, whereas baseline CG is significantly slower. \par


\begin{table}[]
  \centering
  \resizebox{0.9\columnwidth}{!}{
    \begin{tabular}{c|c|c|c|c}
      \toprule[1.5pt]
      \multirow{2}{*}{Method}       & \multirow{2}{*}{Hes.}   & \multirow{2}{*}{\# ($\mathbf{T}$, $\mathbf{V}$)} & \multirow{2}{*}{Setting}                                        & \multirow{2}{*}{\shortstack{Time \\ (Sec.)}}              \\
                                    &                         &                                                  &                                                                 &                                  \\\midrule
      \multirow{3}{*}{\emph{CG}}    & \multirow{3}{*}{\cmark} & ($10^2,10$)                                      & \multirow{3}{*}{\emph{Max\_Iter.} = $\mathbf{T}$}               & 23.1                             \\
                                    &                         & ($10^3,10^2$)                                    &                                                                 & 665.8                            \\
                                    &                         & ($10^4,10^3$)                                    &                                                                 & -                                \\ \midrule
      \multirow{3}{*}{\emph{LISSA}} & \multirow{3}{*}{\cmark} & ($10^2,10$)                                      & \multirow{3}{*}{\shortstack {\emph{Depth} = $\mathbf{T}$ / $10$                                    \\ \emph{Repeat} = $10$}}  &  3.6     \\
                                    &                         & ($10^3,10^2$)                                    &                                                                 & 13.2                             \\
                                    &                         & ($10^4,10^3$)                                    &                                                                 & 136.7                            \\ \midrule
      \multirow{3}{*}{\emph{Hydra}} & \multirow{3}{*}{\xmark} & ($10^2,10$)                                      & \multirow{3}{*}{\emph{Trace\_ID} = $\mathbf{T}$}                & 2.4                     \\
                                    &                         & ($10^3,10^2$)                                    &                                                                 & 6.9                     \\
                                    &                         & ($10^4,10^3$)                                    &                                                                 & 161.1                   \\ \midrule
                                    
      \multirow{3}{*}{\emph{HESIT}} & \multirow{3}{*}{\xmark} & ($10^2,10$)                                      & \multirow{3}{*}{\emph{Trace\_ID} = $\mathbf{T}$}                & \textbf{2.2}                     \\
                                    &                         & ($10^3,10^2$)                                    &                                                                 & \textbf{6.3}                     \\
                                    &                         & ($10^4,10^3$)                                    &                                                                 & \textbf{132.9}                   \\
      \bottomrule[1.5pt]
    \end{tabular}}
  \caption{Computation time for data analysis methods. \# ($\mathbf{T}$, $\mathbf{V}$) denotes that measure the influence of $\mathbf{T}$ training data to the performance on $\mathbf{V}$ validation data. The computing devices are AMD-EPYC 7763 (CPU) and NVIDIA A100 (GPU).}
  \vspace{-0.4cm}
  \label{table5}
\end{table}
\begin{figure}[h!]
  \centering
  \includegraphics[width=0.48\textwidth]{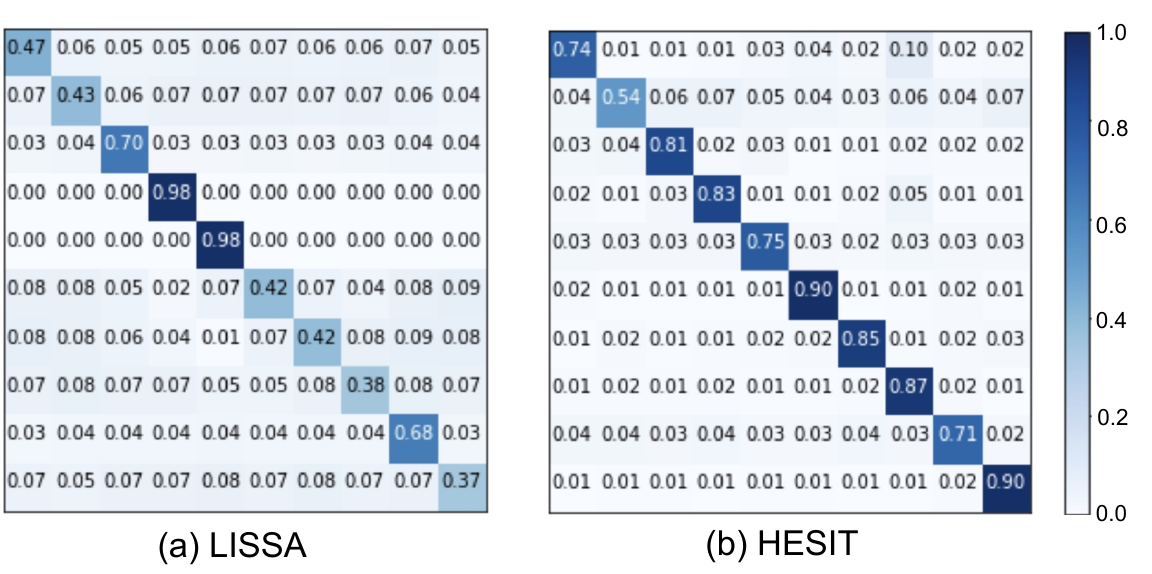}
    \vspace{-0.2cm}
  \caption{Inter-class and Intra-class (diagonal elements) contributions for CIFAR-10 dataset, which are measured by (a) LISSA and (b) HESIT. }
  \label{fig3}
  \vspace{-0.4cm}
\end{figure}

To further demonstrate that HESIT can still properly qualify data contribution even without estimating inverse Hessian, we then visualize the inter-class and intra-class contributions in Fig~\ref{fig3}, where diagonal elements denote the contribution of the intra-class training data to the test class. It is observed that without estimating inverse Hessian, HESIT can still successfully measure the intra-class data influence by hyper-gradient tracing. In addition, it is observed that compared with LISSA, HESIT provides a higher estimation of intra-class contribution due to iteratively tracing.

\section{Conclusion}
In this paper, we propose HESIT to address the catastrophic forgetting in ToDs. It selects exemplar for rehearsal-based CL methods that examines training examples in each domain from a performance perspective. Furthermore, HESIT tracks the hyper-gradient of training examples in an optimization method that is Hessian-free and compatible with large pre-trained models. Experiments show that HESIT effectively overcomes catastrophic forgetting and delivers state-of-the-art performance on the largest ToDs benchmark for CL.
\par

\cleardoublepage

\section*{Limitations} 
When selecting exemplars for the current domain, HESIT only evaluates the performance on the self-domain validation set. However, we suggest that cross-domain influences should be taken into account, which means that the validation set of the later domain in the curriculum might be utilised to examine the picked example of the earlier domain. We would leave it as future work. In addition, we can only compare our model to the state-of-the-art work on ToDs from 2020, as there has been little work on ToDs employing CL in recent years compared to the extensive explorations of CL in computer vision, especially classification task.
\section*{Ethical Consideration}
In developing and evaluating HESIT, our hyper-gradient-based exemplar strategy for Intelligent task-oriented dialogue systems (ToDs), we considered a number of significant ethical implications to ensure that our research conforms to generally accepted standards for the development of responsible AI.

\textbf{Privacy and Data Security}: While HESIT requires the selection of exemplars from the training data, we are mindful of user privacy and data security. The data used for training and testing HESIT are either publicly accessible or anonymized and used with the proper permissions, ensuring that no personally identifiable user information is exploited.

\textbf{Bias and Fairness}: Despite the optimistic results demonstrated by HESIT, it is essential to recognise the inherent possibility of bias in any AI system. We have attempted to mitigate this risk by assuring a diverse and representative training dataset, thereby preventing the model from favouring certain task domains disproportionately over others. Future work will include continued evaluations of potential biases with the objective of developing additional bias mitigation strategies.

\textbf{Transparency and Accountability}: The performance of HESIT is primarily dependent on the exemplars chosen from each task domain. This strategy is effective in terms of model performance, but if not managed carefully, it could contribute to a lack of transparency. We intend to maintain detailed documentation regarding the exemplar selection procedure and to make our algorithms, methods, and results accessible and interpretable to both practitioners and users.

\textbf{Impact on Employment}: The development of ToDs with the capacity for continuous learning may raise concerns regarding potential employment displacement. While our research contributes to the advancement of AI, it is not intended to supplant humans but rather to enhance their capabilities and productivity. To ensure a responsible transition to this technology, we encourage ongoing discussions regarding this issue.

\textbf{Potential Misuse}: As with all AI technologies, there is the potential for misuse with our method. If HESIT is used with nefarious intent or if the model is fed inappropriate data, there is a risk of abuse. We strongly advise establishing appropriate precautions and usage policies.

\textbf{Long-term effects}: We recognise that the long-term effects of HESIT and analogous technologies are indeterminate and require constant monitoring. As researchers, we are committed to continuous monitoring of our technology, its applications, and their societal implications.

In conclusion, we believe that the advantages of developing Continual Learning systems such as HESIT must be weighed against these ethical considerations. As a result, we do not perceive this work as a destination, but rather as a stepping stone on the path towards responsible and beneficial AI.

\bibliography{anthology,custom}

\begin{thebibliography}{50}
\expandafter\ifx\csname natexlab\endcsname\relax\def\natexlab#1{#1}\fi

\bibitem[{Agarwal et~al.(2017)Agarwal, Bullins, and Hazan}]{agarwal2017second}
Naman Agarwal, Brian Bullins, and Elad Hazan. 2017.
\newblock Second-order stochastic optimization for machine learning in linear
  time.
\newblock \emph{The Journal of Machine Learning Research}, 18(1):4148--4187.

\bibitem[{Aljundi et~al.(2019)Aljundi, Lin, Goujaud, and
  Bengio}]{aljundi2019gradient}
Rahaf Aljundi, Min Lin, Baptiste Goujaud, and Yoshua Bengio. 2019.
\newblock Gradient based sample selection for online continual learning.
\newblock \emph{Advances in neural information processing systems}, 32.

\bibitem[{Bae et~al.(2022)Bae, Kwak, Kim, Ham, Kang, Lee, and
  Park}]{bae2022building}
Sanghwan Bae, Donghyun Kwak, Sungdong Kim, Donghoon Ham, Soyoung Kang, Sang-Woo
  Lee, and Woomyoung Park. 2022.
\newblock Building a role specified open-domain dialogue system leveraging
  large-scale language models.
\newblock \emph{arXiv preprint arXiv:2205.00176}.

\bibitem[{Bengio(2000)}]{bengio2000continuous}
Yoshua Bengio. 2000.
\newblock Continuous optimization of hyper-parameters.
\newblock In \emph{Proceedings of the IEEE-INNS-ENNS International Joint
  Conference on Neural Networks. IJCNN 2000. Neural Computing: New Challenges
  and Perspectives for the New Millennium}, volume~1, pages 305--310. IEEE.

\bibitem[{Biesialska et~al.(2020)Biesialska, Biesialska, and
  Costa-Jussa}]{biesialska2020continual}
Magdalena Biesialska, Katarzyna Biesialska, and Marta~R Costa-Jussa. 2020.
\newblock Continual lifelong learning in natural language processing: A survey.
\newblock \emph{arXiv preprint arXiv:2012.09823}.

\bibitem[{Broderick et~al.(2013)Broderick, Boyd, Wibisono, Wilson, and
  Jordan}]{broderick2013streaming}
Tamara Broderick, Nicholas Boyd, Andre Wibisono, Ashia~C Wilson, and Michael~I
  Jordan. 2013.
\newblock Streaming variational bayes.
\newblock \emph{Advances in neural information processing systems}, 26.

\bibitem[{Budzianowski et~al.(2018)Budzianowski, Wen, Tseng, Casanueva, Ultes,
  Ramadan, and Ga{\v{s}}i{\'c}}]{budzianowski2018multiwoz}
Pawe{\l} Budzianowski, Tsung-Hsien Wen, Bo-Hsiang Tseng, Inigo Casanueva,
  Stefan Ultes, Osman Ramadan, and Milica Ga{\v{s}}i{\'c}. 2018.
\newblock Multiwoz--a large-scale multi-domain wizard-of-oz dataset for
  task-oriented dialogue modelling.
\newblock \emph{arXiv preprint arXiv:1810.00278}.

\bibitem[{Byrne et~al.(2019)Byrne, Krishnamoorthi, Sankar, Neelakantan,
  Duckworth, Yavuz, Goodrich, Dubey, Cedilnik, and Kim}]{byrne2019taskmaster}
Bill Byrne, Karthik Krishnamoorthi, Chinnadhurai Sankar, Arvind Neelakantan,
  Daniel Duckworth, Semih Yavuz, Ben Goodrich, Amit Dubey, Andy Cedilnik, and
  Kyu-Young Kim. 2019.
\newblock Taskmaster-1: Toward a realistic and diverse dialog dataset.
\newblock \emph{arXiv preprint arXiv:1909.05358}.

\bibitem[{Chaudhry et~al.(2018)Chaudhry, Ranzato, Rohrbach, and
  Elhoseiny}]{chaudhry2018efficient}
Arslan Chaudhry, Marc'Aurelio Ranzato, Marcus Rohrbach, and Mohamed Elhoseiny.
  2018.
\newblock Efficient lifelong learning with a-gem.
\newblock \emph{arXiv preprint arXiv:1812.00420}.

\bibitem[{Chaudhry et~al.(2019)Chaudhry, Rohrbach, Elhoseiny, Ajanthan,
  Dokania, Torr, and Ranzato}]{chaudhry2019continual}
Arslan Chaudhry, Marcus Rohrbach, Mohamed Elhoseiny, Thalaiyasingam Ajanthan,
  Puneet~K Dokania, Philip~HS Torr, and M~Ranzato. 2019.
\newblock Continual learning with tiny episodic memories.
\newblock \emph{arXiv preprint arXiv:1902.10486}.

\bibitem[{Chen et~al.(2021)Chen, Li, Yu, Wu, and Miao}]{chen2021hydra}
Yuanyuan Chen, Boyang Li, Han Yu, Pengcheng Wu, and Chunyan Miao. 2021.
\newblock Hydra: Hypergradient data relevance analysis for interpreting deep
  neural networks.
\newblock In \emph{Proceedings of the AAAI Conference on Artificial
  Intelligence}, volume~35, pages 7081--7089.

\bibitem[{Cui et~al.(2019)Cui, Jia, Lin, Song, and Belongie}]{cui2019class}
Yin Cui, Menglin Jia, Tsung-Yi Lin, Yang Song, and Serge Belongie. 2019.
\newblock Class-balanced loss based on effective number of samples.
\newblock In \emph{Proceedings of the IEEE/CVF conference on computer vision
  and pattern recognition}, pages 9268--9277.

\bibitem[{De~Lange et~al.(2021)De~Lange, Aljundi, Masana, Parisot, Jia,
  Leonardis, Slabaugh, and Tuytelaars}]{de2021continual}
Matthias De~Lange, Rahaf Aljundi, Marc Masana, Sarah Parisot, Xu~Jia,
  Ale{\v{s}} Leonardis, Gregory Slabaugh, and Tinne Tuytelaars. 2021.
\newblock A continual learning survey: Defying forgetting in classification
  tasks.
\newblock \emph{IEEE transactions on pattern analysis and machine
  intelligence}, 44(7):3366--3385.

\bibitem[{Ermis et~al.(2022)Ermis, Zappella, Wistuba, Rawal, and
  Archambeau}]{ermis2022continual}
Beyza Ermis, Giovanni Zappella, Martin Wistuba, Aditya Rawal, and C{\'e}dric
  Archambeau. 2022.
\newblock Continual learning with transformers for image classification.
\newblock In \emph{Proceedings of the IEEE/CVF Conference on Computer Vision
  and Pattern Recognition}, pages 3774--3781.

\bibitem[{Guo et~al.(2020)Guo, Rajani, Hase, Bansal, and Xiong}]{guo2020fastif}
Han Guo, Nazneen~Fatema Rajani, Peter Hase, Mohit Bansal, and Caiming Xiong.
  2020.
\newblock Fastif: Scalable influence functions for efficient model
  interpretation and debugging.
\newblock \emph{arXiv preprint arXiv:2012.15781}.

\bibitem[{Huang et~al.(2021)Huang, Zhang, Chen, Wang, and
  Yang}]{huang2021continual}
Yufan Huang, Yanzhe Zhang, Jiaao Chen, Xuezhi Wang, and Diyi Yang. 2021.
\newblock Continual learning for text classification with information
  disentanglement based regularization.
\newblock \emph{arXiv preprint arXiv:2104.05489}.

\bibitem[{Isele and Cosgun(2018)}]{isele2018selective}
David Isele and Akansel Cosgun. 2018.
\newblock Selective experience replay for lifelong learning.
\newblock In \emph{Proceedings of the AAAI Conference on Artificial
  Intelligence}, volume~32.

\bibitem[{Kale and Rastogi(2020)}]{kale2020few}
Mihir Kale and Abhinav Rastogi. 2020.
\newblock Few-shot natural language generation by rewriting templates.
\newblock \emph{arXiv preprint arXiv:2004.15006}.

\bibitem[{Kirkpatrick et~al.(2017)Kirkpatrick, Pascanu, Rabinowitz, Veness,
  Desjardins, Rusu, Milan, Quan, Ramalho, Grabska-Barwinska
  et~al.}]{kirkpatrick2017overcoming}
James Kirkpatrick, Razvan Pascanu, Neil Rabinowitz, Joel Veness, Guillaume
  Desjardins, Andrei~A Rusu, Kieran Milan, John Quan, Tiago Ramalho, Agnieszka
  Grabska-Barwinska, et~al. 2017.
\newblock Overcoming catastrophic forgetting in neural networks.
\newblock \emph{Proceedings of the national academy of sciences},
  114(13):3521--3526.

\bibitem[{Koh and Liang(2017)}]{koh2017understanding}
Pang~Wei Koh and Percy Liang. 2017.
\newblock Understanding black-box predictions via influence functions.
\newblock In \emph{International conference on machine learning}, pages
  1885--1894. PMLR.

\bibitem[{Krizhevsky et~al.(2009)Krizhevsky, Hinton
  et~al.}]{krizhevsky2009learning}
Alex Krizhevsky, Geoffrey Hinton, et~al. 2009.
\newblock Learning multiple layers of features from tiny images.

\bibitem[{Lee(2017)}]{lee2017toward}
Sungjin Lee. 2017.
\newblock Toward continual learning for conversational agents.
\newblock \emph{arXiv preprint arXiv:1712.09943}.

\bibitem[{Li et~al.(2019)Li, Zhou, Wu, Socher, and Xiong}]{li2019learn}
Xilai Li, Yingbo Zhou, Tianfu Wu, Richard Socher, and Caiming Xiong. 2019.
\newblock Learn to grow: A continual structure learning framework for
  overcoming catastrophic forgetting.
\newblock In \emph{International Conference on Machine Learning}, pages
  3925--3934. PMLR.

\bibitem[{Lin et~al.(2021)Lin, Madotto, Winata, Xu, Jiang, Hu, Shi, and
  Fung}]{lin2021bitod}
Zhaojiang Lin, Andrea Madotto, Genta~Indra Winata, Peng Xu, Feijun Jiang,
  Yuxiang Hu, Chen Shi, and Pascale Fung. 2021.
\newblock Bitod: A bilingual multi-domain dataset for task-oriented dialogue
  modeling.
\newblock \emph{arXiv preprint arXiv:2106.02787}.

\bibitem[{Madotto et~al.(2020)Madotto, Lin, Zhou, Moon, Crook, Liu, Yu, Cho,
  and Wang}]{madotto2020continual}
Andrea Madotto, Zhaojiang Lin, Zhenpeng Zhou, Seungwhan Moon, Paul Crook, Bing
  Liu, Zhou Yu, Eunjoon Cho, and Zhiguang Wang. 2020.
\newblock Continual learning in task-oriented dialogue systems.
\newblock \emph{arXiv preprint arXiv:2012.15504}.

\bibitem[{Mai et~al.(2022)Mai, Li, Jeong, Quispe, Kim, and
  Sanner}]{mai2022online}
Zheda Mai, Ruiwen Li, Jihwan Jeong, David Quispe, Hyunwoo Kim, and Scott
  Sanner. 2022.
\newblock Online continual learning in image classification: An empirical
  survey.
\newblock \emph{Neurocomputing}, 469:28--51.

\bibitem[{Maltoni and Lomonaco(2019)}]{maltoni2019continuous}
Davide Maltoni and Vincenzo Lomonaco. 2019.
\newblock Continuous learning in single-incremental-task scenarios.
\newblock \emph{Neural Networks}, 116:56--73.

\bibitem[{Martens et~al.(2010)}]{martens2010deep}
James Martens et~al. 2010.
\newblock Deep learning via hessian-free optimization.
\newblock In \emph{ICML}, volume~27, pages 735--742.

\bibitem[{McCloskey and Cohen(1989)}]{mccloskey1989catastrophic}
Michael McCloskey and Neal~J Cohen. 1989.
\newblock Catastrophic interference in connectionist networks: The sequential
  learning problem.
\newblock In \emph{Psychology of learning and motivation}, volume~24, pages
  109--165. Elsevier.

\bibitem[{Mi et~al.(2020)Mi, Chen, Zhao, Huang, and Faltings}]{mi2020continual}
Fei Mi, Liangwei Chen, Mengjie Zhao, Minlie Huang, and Boi Faltings. 2020.
\newblock Continual learning for natural language generation in task-oriented
  dialog systems.
\newblock \emph{arXiv preprint arXiv:2010.00910}.

\bibitem[{Mundt et~al.(2020)Mundt, Hong, Pliushch, and
  Ramesh}]{mundt2020wholistic}
Martin Mundt, Yong~Won Hong, Iuliia Pliushch, and Visvanathan Ramesh. 2020.
\newblock A wholistic view of continual learning with deep neural networks:
  Forgotten lessons and the bridge to active and open world learning.
\newblock \emph{arXiv preprint arXiv:2009.01797}.

\bibitem[{Papineni et~al.(2002)Papineni, Roukos, Ward, and
  Zhu}]{papineni2002bleu}
Kishore Papineni, Salim Roukos, Todd Ward, and Wei-Jing Zhu. 2002.
\newblock Bleu: a method for automatic evaluation of machine translation.
\newblock In \emph{Proceedings of the 40th annual meeting of the Association
  for Computational Linguistics}, pages 311--318.

\bibitem[{Pruthi et~al.(2020)Pruthi, Liu, Kale, and
  Sundararajan}]{pruthi2020estimating}
Garima Pruthi, Frederick Liu, Satyen Kale, and Mukund Sundararajan. 2020.
\newblock Estimating training data influence by tracing gradient descent.
\newblock \emph{Advances in Neural Information Processing Systems},
  33:19920--19930.

\bibitem[{Qin et~al.(2023)Qin, Chen, and Joty}]{qin-etal-2023-lifelong}
Chengwei Qin, Chen Chen, and Shafiq Joty. 2023.
\newblock \href {https://doi.org/10.18653/v1/2023.emnlp-main.414} {Lifelong
  sequence generation with dynamic module expansion and adaptation}.
\newblock In \emph{Proceedings of the 2023 Conference on Empirical Methods in
  Natural Language Processing}, pages 6701--6714, Singapore. Association for
  Computational Linguistics.

\bibitem[{Qin et~al.(2024)Qin, Chen, Zhao, Xia, and Joty}]{qin2024lifelong}
Chengwei Qin, Ruirui Chen, Ruochen Zhao, Wenhan Xia, and Shafiq Joty. 2024.
\newblock \href {https://openreview.net/forum?id=0DUT9sssWQ} {Lifelong event
  detection with embedding space separation and compaction}.
\newblock In \emph{2024 Annual Conference of the North American Chapter of the
  Association for Computational Linguistics}.

\bibitem[{Qin and Joty(2022{\natexlab{a}})}]{qin-joty-2022-continual}
Chengwei Qin and Shafiq Joty. 2022{\natexlab{a}}.
\newblock \href {https://doi.org/10.18653/v1/2022.acl-long.198} {Continual
  few-shot relation learning via embedding space regularization and data
  augmentation}.
\newblock In \emph{Proceedings of the 60th Annual Meeting of the Association
  for Computational Linguistics (Volume 1: Long Papers)}, pages 2776--2789,
  Dublin, Ireland. Association for Computational Linguistics.

\bibitem[{Qin and Joty(2022{\natexlab{b}})}]{qin2022lfpt}
Chengwei Qin and Shafiq Joty. 2022{\natexlab{b}}.
\newblock \href {https://openreview.net/forum?id=HCRVf71PMF} {{LFPT}5: A
  unified framework for lifelong few-shot language learning based on prompt
  tuning of t5}.
\newblock In \emph{International Conference on Learning Representations}.

\bibitem[{Qu et~al.(2021)Qu, Rahmani, Xu, Williams, and Liu}]{qu2021recent}
Haoxuan Qu, Hossein Rahmani, Li~Xu, Bryan Williams, and Jun Liu. 2021.
\newblock Recent advances of continual learning in computer vision: An
  overview.
\newblock \emph{arXiv preprint arXiv:2109.11369}.

\bibitem[{Radford et~al.(2019)Radford, Wu, Child, Luan, Amodei, Sutskever
  et~al.}]{radford2019language}
Alec Radford, Jeffrey Wu, Rewon Child, David Luan, Dario Amodei, Ilya
  Sutskever, et~al. 2019.
\newblock Language models are unsupervised multitask learners.
\newblock \emph{OpenAI blog}, 1(8):9.

\bibitem[{Rebuffi et~al.(2017)Rebuffi, Kolesnikov, Sperl, and
  Lampert}]{rebuffi2017icarl}
Sylvestre-Alvise Rebuffi, Alexander Kolesnikov, Georg Sperl, and Christoph~H
  Lampert. 2017.
\newblock icarl: Incremental classifier and representation learning.
\newblock In \emph{Proceedings of the IEEE conference on Computer Vision and
  Pattern Recognition}, pages 2001--2010.

\bibitem[{Rolnick et~al.(2019)Rolnick, Ahuja, Schwarz, Lillicrap, and
  Wayne}]{rolnick2019experience}
David Rolnick, Arun Ahuja, Jonathan Schwarz, Timothy Lillicrap, and Gregory
  Wayne. 2019.
\newblock Experience replay for continual learning.
\newblock \emph{Advances in Neural Information Processing Systems}, 32.

\bibitem[{Rusu et~al.(2016)Rusu, Rabinowitz, Desjardins, Soyer, Kirkpatrick,
  Kavukcuoglu, Pascanu, and Hadsell}]{rusu2016progressive}
Andrei~A Rusu, Neil~C Rabinowitz, Guillaume Desjardins, Hubert Soyer, James
  Kirkpatrick, Koray Kavukcuoglu, Razvan Pascanu, and Raia Hadsell. 2016.
\newblock Progressive neural networks.
\newblock \emph{arXiv preprint arXiv:1606.04671}.

\bibitem[{Serra et~al.(2018)Serra, Suris, Miron, and
  Karatzoglou}]{serra2018overcoming}
Joan Serra, Didac Suris, Marius Miron, and Alexandros Karatzoglou. 2018.
\newblock Overcoming catastrophic forgetting with hard attention to the task.
\newblock In \emph{International Conference on Machine Learning}, pages
  4548--4557. PMLR.

\bibitem[{Sun et~al.(2019)Sun, Ho, and Lee}]{sun2019lamol}
Fan-Keng Sun, Cheng-Hao Ho, and Hung-Yi Lee. 2019.
\newblock Lamol: Language modeling for lifelong language learning.
\newblock \emph{arXiv preprint arXiv:1909.03329}.

\bibitem[{Sun et~al.(2022)Sun, Lyu, Shang, Feng, and Wan}]{sun2022exploring}
Qing Sun, Fan Lyu, Fanhua Shang, Wei Feng, and Liang Wan. 2022.
\newblock Exploring example influence in continual learning.
\newblock \emph{arXiv preprint arXiv:2209.12241}.

\bibitem[{Verwimp et~al.(2021)Verwimp, De~Lange, and
  Tuytelaars}]{verwimp2021rehearsal}
Eli Verwimp, Matthias De~Lange, and Tinne Tuytelaars. 2021.
\newblock Rehearsal revealed: The limits and merits of revisiting samples in
  continual learning.
\newblock In \emph{Proceedings of the IEEE/CVF International Conference on
  Computer Vision}, pages 9385--9394.

\bibitem[{Wen et~al.(2015)Wen, Gasic, Mrksic, Su, Vandyke, and
  Young}]{wen2015semantically}
Tsung-Hsien Wen, Milica Gasic, Nikola Mrksic, Pei-Hao Su, David Vandyke, and
  Steve Young. 2015.
\newblock Semantically conditioned lstm-based natural language generation for
  spoken dialogue systems.
\newblock \emph{arXiv preprint arXiv:1508.01745}.

\bibitem[{Wortsman et~al.(2020)Wortsman, Ramanujan, Liu, Kembhavi, Rastegari,
  Yosinski, and Farhadi}]{wortsman2020supermasks}
Mitchell Wortsman, Vivek Ramanujan, Rosanne Liu, Aniruddha Kembhavi, Mohammad
  Rastegari, Jason Yosinski, and Ali Farhadi. 2020.
\newblock Supermasks in superposition.
\newblock \emph{Advances in Neural Information Processing Systems},
  33:15173--15184.

\bibitem[{Wu et~al.(2019)Wu, Madotto, Hosseini-Asl, Xiong, Socher, and
  Fung}]{wu2019transferable}
Chien-Sheng Wu, Andrea Madotto, Ehsan Hosseini-Asl, Caiming Xiong, Richard
  Socher, and Pascale Fung. 2019.
\newblock Transferable multi-domain state generator for task-oriented dialogue
  systems.
\newblock \emph{arXiv preprint arXiv:1905.08743}.

\bibitem[{Xu and Zhu(2018)}]{xu2018reinforced}
Ju~Xu and Zhanxing Zhu. 2018.
\newblock Reinforced continual learning.
\newblock \emph{Advances in Neural Information Processing Systems}, 31.

\end{thebibliography}
\bibliographystyle{acl_natbib}

\newpage
\appendix
\section{Large Language Model and ToDs}
Large Language Models (LLMs) have emerged as an epistemic beacon in the field of natural language processing (NLP), endowing text-based tasks with substantial performance gains including ToDs~\cite{bae2022building}. Furthermore, a foundation model with multiple adapters seems to be potential solution for overcoming catastrophic forgetting. However, this combination can not fit continual learning in ToDs, especially with increasing task domains. Firstly, the task domain of inference data is agnostic that requires a extra procedure to determine which of adapter should be activated for inference. Secondly, no matter what procedure is utilized, all of 37 combinations need to be traversed once, resulting in unacceptable delay for ToDs. Therefore, instead of using LLMs with adapter, we utilize single model to adapt to all domains. 

\section{Examples of E2E ToDs}
\label{data_format}
\begin{figure}[]
  \centering
  \includegraphics[width=0.48\textwidth]{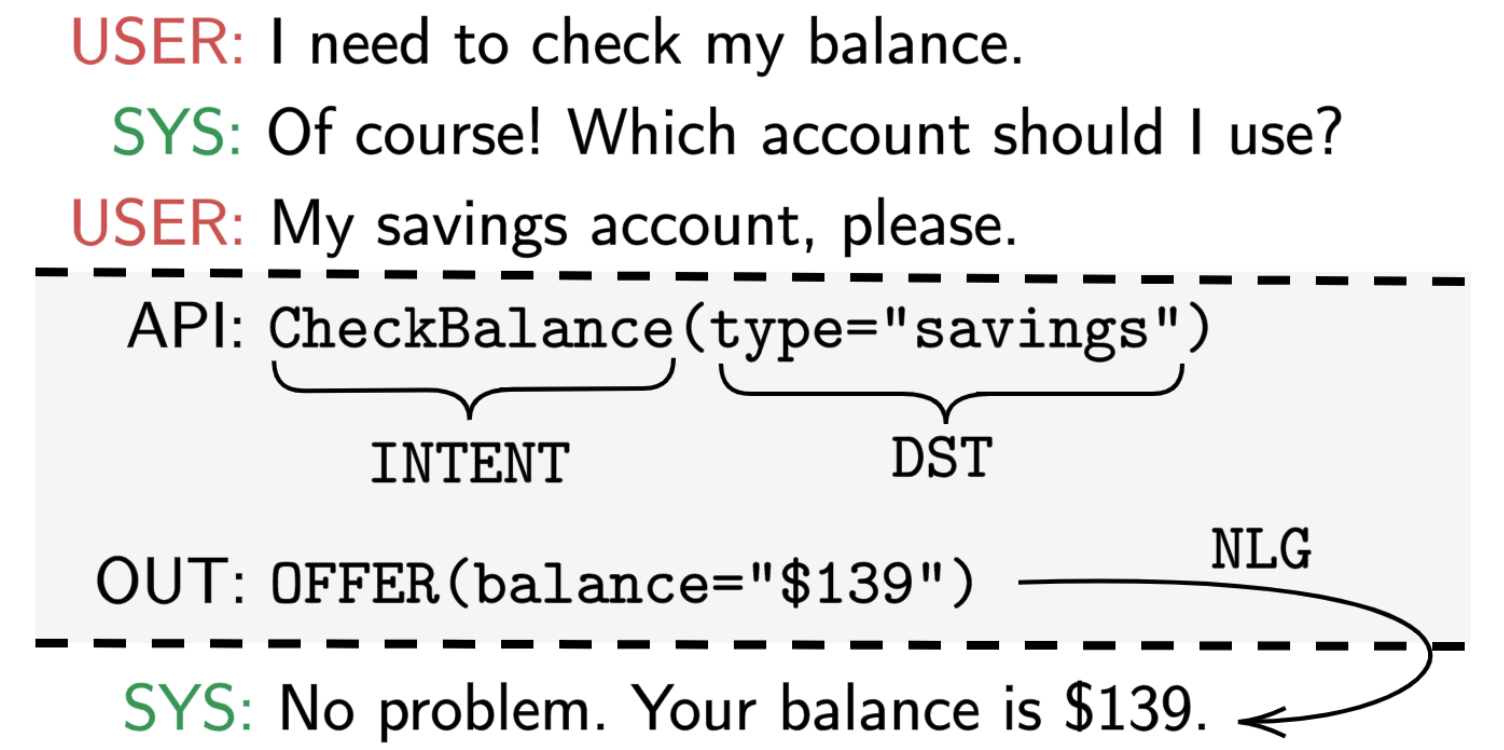}
  \caption{Example of input-out pairs, for the four settings, INTENT, DST, NLG and end-to-end (E2E).}
  \label{fig2}
\end{figure}

We first give an example to elaborate INTENT, DST, NLG, and E2E. As shown in Fig~\ref{fig2}, the grey box denotes the API-call $\mathcal{C}(\mathbf{H})$ process, where the user intents (INTENT) and slot-value pairs (DST) are both predicted according to dialogue history $\mathbf{H}$. Then the system will generate a response (NLG) based on both $\mathbf{H}$ and $\mathcal{C}(\mathbf{H})$: $\mathbf{H}+\mathcal{C}(\mathbf{H})\rightarrow\mathbf{R}$. In addition, when $\mathcal{C}(\mathbf{H})$ is empty, the system can directly generate responses based on history: $\mathbf{H} \rightarrow \mathbf{R}$. \par

\section{Methodology Supplement}
\label{method_sub}
For each task domain, Let $\mathcal{D}_t$ consists of training set $\mathcal{D}_t^\textrm{trn}$, validation set $\mathcal{D}_t^\textrm{val}$, and test set $\mathcal{D}_t^{tst}$. Assume $\mathcal{D}_t^\textrm{trn}$ contains $N$ training-points which is defined as $z_i=(x_i, y_i)$. Our goal is to pick out those influential $z_i$s that benefit model performance on unseen test-point $z'\in \mathcal{D}_t^\textrm{val}$, as $\mathcal{D}_t^{tst}$ is unavailable during training. Leveraging derivation chain rule, the IFs~\cite{koh2017understanding} define the influence of $z$ on $z'$ as:
\begin{equation}
  \mathcal{I}(z,z')= -\nabla_{\theta}\ell(z',\hat{\theta})^TH^{-1}_{\hat{\theta}}\nabla_{\theta}\ell(z,\hat{\theta})
\end{equation}
where $H_{\hat{\theta}}=\nabla_{\theta}^2\ell(z,\hat{\theta})$ is the Hessian and is positive definite (PD) by assumption. In practice, $H_{\hat{\theta}}$ can not be materialized in memory, let alone be inverted by standard linear algebra. \par
Mainstream methods employ LISSA~\cite{agarwal2017second}, which only samples a single point per iteration as an estimator of inverse Hessian. Compared with the standard transformation of matrix inversion~\cite{martens2010deep}, LISSA shows significant speedups in the following iteration $r$:
\begin{equation}
  H_r^{-1}v=v+(1-H)H^{-1}_{r-1}v
\end{equation}
where $H$ is approximated on random batch and $v=\nabla_{\theta}\ell(z',\hat{\theta})$ is a gradient. However, this method faces two drawbacks. Firstly, only the final parameters are utilized to calculate the gradient, while the data influence is involved in a dynamic optimization process. Secondly, Hessian might not be positive definite for a large pre-trained model, and the estimation is not accurate enough. In practice, we found that using LISSA to estimate the inverse Hessian of GPT-2 is unstable and time-consuming. \par
To address the above defects, \emph{TracIn}~\cite{pruthi2020estimating} traces the changes in the loss across all gradient steps, and avoids the estimation of inverse Hessian by the following definition:
\begin{equation}
  \mathcal{I}_{\emph{TracIn}}(z,z')= \frac{1}{C} \sum_{i=1}^C \gamma_i \ \nabla_{\theta_i} \ell(z,\hat{\theta_i}) \cdot \nabla_{\theta_i} \ell(z',\hat{\theta_i})
\end{equation}
where $C$ is the number of the checkpoints during training. Despite considering the optimization process, the computing complexity is $C$ times slower than that of using exact gradient similarity and, as discussed in~\cite{pruthi2020estimating}, care needs to be taken in selecting checkpoints. Another practical obstacle of \emph{TracIn} in CL settings is that only several epochs are trained for each domain, resulting in insufficient $C$ for influence analysis.

\section{Dataset Statistics}
\label{dataset}
The main dataset statistics are shown in Table~\ref{table6}. It contains 37 domains, 280 types of intents, and more than 31K training dialogues. The detailed example amounts of modules as well as training, validation, and test set are illustrated in Table~\ref{table7}. \par
\begin{table}[h]
  \resizebox{\linewidth}{!}{
    \begin{tabular}{r|ccc|ccc}
      \toprule[1.5pt]
      \textbf{Name} & \textbf{Train} & \textbf{Valid} & \textbf{Test} & \textbf{Dom.} & \textbf{Intents} & \textbf{Turns} \\ \midrule
      TM19          & 4,403          & 551            & 553           & 6             & 112              & 19.97          \\
      TM20          & 13,839         & 1,731          & 1,734         & 7             & 128              & 16.92          \\
      MWoZ          & 7,906          & 1,000          & 1,000         & 5             & 15               & 13.93          \\
      SGD           & 5,278          & 761            & 1,531         & 19            & 43               & 14.71          \\ \midrule
      Total         & 31,426         & 4,043          & 4,818         & 37            & 280              & 16.23          \\ \bottomrule[1.5pt]
    \end{tabular}
  }
  \caption{Main datasets statistics. }
  \label{table6}
\end{table}

Due to a serious imbalance of task domain, we observe that some CL methods (e.g., EWC) are vulnerable to the task order in the curriculum. If the model first learns on a small domain, then the regularization item would influence the subsequent learning on large domains. Meanwhile, we found the data replay methods are more insensitive to task order, as the data amount of each domain stored for retraining is equal. In addition, we have also attempted to sample different amounts of exemplars according to the size of the domain. However, it achieves similar results compared with sampling an equal amount of exemplars for each domain.

\begin{table*}[t]
  \begin{tabular}{r|ccc|ccc|ccc}
    \toprule[1.5pt]
    \multirow{2}{*}{\textbf{Domains}}  & \multicolumn{3}{c}{\textbf{DST \& INTENT}} & \multicolumn{3}{c}{\textbf{NLG}} & \multicolumn{3}{c}{\textbf{End-to-End}}                                                                                                 \\
                                       & \textit{Train}                             & \textit{Dev}                     & \textit{Test}                           & \textit{Train} & \textit{Dev} & \textit{Test} & \textit{Train} & \textit{Dev} & \textit{Test} \\ \midrule[1.5pt]
    {\ul \textit{TM19 movie}}          & 4733                                       & 584                              & 500                                     & 3010           & 366          & 341           & 12766          & 1632         & 1481          \\
    {\ul \textit{TM19 auto}}           & 3897                                       & 448                              & 522                                     & 2128           & 223          & 283           & 10918          & 1248         & 1443          \\
    {\ul \textit{TM19 restaurant}}     & 4434                                       & 568                              & 561                                     & 2582           & 330          & 333           & 12862          & 1669         & 1630          \\
    {\ul \textit{TM19 pizza}}          & 2883                                       & 381                              & 359                                     & 1326           & 171          & 171           & 8720           & 1145         & 1083          \\
    {\ul \textit{TM19 uber}}           & 4378                                       & 535                              & 525                                     & 2418           & 290          & 278           & 11331          & 1362         & 1361          \\
    {\ul \textit{TM19 coffee}}         & 2591                                       & 302                              & 335                                     & 1381           & 151          & 184           & 7429           & 894          & 936           \\
    {\ul \textit{TM20 flight}}         & 15868                                      & 1974                             & 1940                                    & 10148          & 1272         & 1245          & 36778          & 4579         & 4569          \\
    {\ul \textit{TM20 food-ordering}}  & 3404                                       & 411                              & 431                                     & 2394           & 277          & 287           & 7838           & 941          & 986           \\
    {\ul \textit{TM20 hotel}}          & 15029                                      & 1908                             & 1960                                    & 6590           & 842          & 869           & 35022          & 4400         & 4532          \\
    {\ul \textit{TM20 music}}          & 5917                                       & 764                              & 769                                     & 4196           & 537          & 523           & 13723          & 1773         & 1787          \\
    {\ul \textit{TM20 restaurant}}     & 13738                                      & 1761                             & 1691                                    & 8356           & 1063         & 994           & 34560          & 4398         & 4297          \\
    {\ul \textit{TM20 sport}}          & 13072                                      & 1668                             & 1654                                    & 12044          & 1553         & 1542          & 29391          & 3765         & 3723          \\
    {\ul \textit{TM20 movie}}          & 13221                                      & 1703                             & 1567                                    & 9406           & 1203         & 1093          & 32423          & 4158         & 3881          \\
    {\ul \textit{MWOZ taxi}}           & 1239                                       & 234                              & 194                                     & 402            & 71           & 56            & 2478           & 468          & 388           \\
    {\ul \textit{MWOZ train}}          & 1452                                       & 158                              & 160                                     & 563            & 63           & 59            & 2905           & 316          & 320           \\
    {\ul \textit{MWOZ restaurant}}     & 5227                                       & 243                              & 281                                     & 3333           & 141          & 177           & 10461          & 486          & 563           \\
    {\ul \textit{MWOZ hotel}}          & 2798                                       & 289                              & 385                                     & 1924           & 194          & 258           & 5602           & 579          & 771           \\
    {\ul \textit{MWOZ attraction}}     & 484                                        & 43                               & 42                                      & 295            & 27           & 26            & 975            & 86           & 85            \\
    {\ul \textit{sgd restaurants}}     & 2686                                       & 278                              & 616                                     & 1720           & 166          & 386           & 5756           & 606          & 1354          \\
    {\ul \textit{sgd media}}           & 1411                                       & 230                              & 458                                     & 988            & 167          & 324           & 3114           & 502          & 1005          \\
    {\ul \textit{sgd events}}          & 4881                                       & 598                              & 989                                     & 3241           & 389          & 590           & 10555          & 1317         & 2197          \\
    {\ul \textit{sgd music}}           & 1892                                       & 275                              & 556                                     & 1506           & 224          & 464           & 4040           & 597          & 1215          \\
    {\ul \textit{sgd movies}}          & 1665                                       & 181                              & 52                                      & 996            & 114          & 44            & 3760           & 420          & 126           \\
    {\ul \textit{sgd flights}}         & 4766                                       & 1041                             & 1756                                    & 2571           & 627          & 982           & 10429          & 2244         & 3833          \\
    {\ul \textit{sgd ridesharing}}     & 652                                        & 85                               & 187                                     & 377            & 48           & 107           & 1448           & 188          & 418           \\
    {\ul \textit{sgd rentalcars}}      & 1510                                       & 250                              & 469                                     & 865            & 153          & 280           & 3277           & 538          & 1009          \\
    {\ul \textit{sgd buses}}           & 1862                                       & 331                              & 653                                     & 1102           & 218          & 412           & 4050           & 709          & 1393          \\
    {\ul \textit{sgd hotels}}          & 3237                                       & 394                              & 948                                     & 1997           & 243          & 597           & 6983           & 858          & 2053          \\
    {\ul \textit{sgd services}}        & 3328                                       & 360                              & 926                                     & 2225           & 230          & 611           & 7262           & 803          & 2016          \\
    {\ul \textit{sgd homes}}           & 2098                                       & 170                              & 533                                     & 1312           & 96           & 338           & 4519           & 394          & 1158          \\
    {\ul \textit{sgd banks}}           & 1188                                       & 139                              & 293                                     & 723            & 84           & 181           & 2599           & 319          & 667           \\
    {\ul \textit{sgd calendar}}        & 592                                        & 115                              & 236                                     & 397            & 65           & 133           & 1313           & 246          & 501           \\
    {\ul \textit{sgd alarm}}           & 212                                        & 34                               & 91                                      & 221            & 30           & 74            & 580            & 82           & 198           \\
    {\ul \textit{sgd weather}}         & 196                                        & 32                               & 80                                      & 123            & 23           & 59            & 433            & 70           & 169           \\
    {\ul \textit{sgd travel}}          & 186                                        & 23                               & 48                                      & 121            & 14           & 30            & 420            & 53           & 106           \\
    {\ul \textit{sgd payment}}         & 227                                        & 21                               & 51                                      & 143            & 14           & 32            & 497            & 44           & 113           \\
    {\ul \textit{sgd trains}}          & 300                                        & 73                               & 128                                     & 149            & 43           & 66            & 668            & 158          & 274           \\ \midrule
    \multicolumn{1}{r}{\textbf{Total}} & 147254                                     & 18604                            & 22946                                   & 93273          & 11722        & 14429         & 347885         & 44047        & 53641         \\ \bottomrule[1.5pt]
  \end{tabular}
  \caption{Detained data amounts for each domian and module}
  \label{table7}
\end{table*}

\section{Experimental Details}
\label{cifar}
\subsection{CIFAR-10 Dataset}
The CIFAR-10 dataset~\cite{krizhevsky2009learning} consists of 60000 32x32 color images in 10 classes, which can be split into 50000 training images and 10000 test images. There are 10 different classes in the CIFAR-10 dataset, including cars, birds, cats, deer, dogs, frogs, horses, ships, and trucks. Each class contains 6000 images.\par
\subsection{CNN backbone}
We employ a simple convolutional neural network, including 2$\times$2D CNN layers with pooling layers, then a ReLU activation function and 2$\times$ linear layer are added to predict the labels. \par
In order to obtain Fig.~\ref{fig3}, the CNN network is trained by an SGD optimizer with a momentum of 0.9. The learning rate is $1 \times e^{-3}$ and 50 training epochs are repeated. Then we calculate the contribution of training data to test data regarding different categories.
\end{document}